\documentclass[twocolumn]{article}

\usepackage{amsmath,amssymb,amsthm}
\usepackage{tabularx,adjustbox}

\title{A Multi-Modal Feature Embedding Approach to Diagnose Alzheimer Disease from Spoken Language}

\author{
Haj Zargarbashi, S. Soroush \\
\texttt{s.zargarbashi@ut.ac.ir}
\and
Babaali, Bagher \\
\texttt{babaali@ut.ac.ir}
}

\begin{document}
	\maketitle
	
	\begin{abstract}
		Introduction: Alzheimer disease is a type of dementia in which the early diagnosis plays a major rule in the quality of treatment. Among new works in diagnosis of Alzheimer disease, there are many of them analyzing the voice stream acoustically, syntactically or both. The mostly used tools to perform these analysis usually include machine learning techniques.
		
		Objective: Designing an automatic machine learning based diagnosis system will help in the procedure of early detection. Also systems, using noninvasive data are preferable.
		
		Methods: We used are classification system based on spoken language. We use three (statistical and neural) approaches to classify audio signals from spoken language into two classes of dementia and control.
		
		Result: This work designs a multi-modal feature embedding on the spoken language audio signal using three approaches; $N$-gram, i-vector and x-vector. The evaluation of the system is done on cookie picture description task from Pitt Corpus dementia bank with accuracy of $83.6\%$.

%
	
	\end{abstract}

	\section{Introduction}\label{sec:Introduction}

	Alzheimer (AD) is a neurodegenerative disease mostly targeting aged part of society, and also a cause of 60 to 80 percent of dementia cases (See more about dementia in \cite{ritchie2002dementias}). Its early symptoms are including difficulty in remembering recent conversations, names or events, and by passing time, can lead to impaired communication, disorientation, confusion, poor judgment, behavioral changes, etc. These disorders are results of damages and destruction of neurons involved in cognitive functions. Procedure of Alzheimer's disease can lead to problems in many basic functionalities such as walking and swallowing in which in these cases patients should be under full time protection and clinical care \cite{alzheimer20182018}. Some observations on symptoms like impaired awareness in AD are studied in \cite{rymer2002impaired}.  As long as AD causes are unknown, there are no certain protocols to prevent it \cite{daviglus2010national}. 
	
	There is a wide area of research on AD diagnosis both in clinical (medical and psychological) and computational area. Although there is no any specific test to diagnose AD, there are some work-flows used by physicians (with aid of specialists) to help make a diagnosis. These procedures usually involve medical and family history, tracking changes in thinking skills and behaviors, blood tests and brain imaging (which is used usually to find high level marks of Alzheimer's) \cite{alzheimer20182018}. Usually an Alzheimer case is verified via mental status test (evaluation on memory, ability to solve simple problems and other thinking skills), neurological exam, medical history, etc.
	
	The procedure of Alzheimer's treatment is hard since non of pharmacological treatments used, stop or even slow the damage and destruction of neurons, and some of them improve the symptoms by increasing chemicals called neurotransmitters in brain. On the other hand non-pharmacological therapies are aimed to maintain and improve cognitive functions and they do not slow or stop the neurons destruction too \cite{alzheimer20182018}.
	
	Early diagnosis of AD, as long as it helps to begin the treatment process earlier in which it helps to optimize patient's functions by temporarily improving and prolong cognitive function, is considered very helpful medically. Also on the other hand for the patients, it provides more time to make important decisions with more cognitive abilities. Because of the benefits that early diagnosis has, there is a vast research in diagnosis of AD through understanding biomarkers and other methods.

	The usual procedure of AD detection takes days or sometimes weeks to complete and due to its length and cost is not widely common. and as discussed early AD detection can be helpful for preserving patient's cognitive functions and also provides better chance of benefiting from treatment. The early (and usually non invasive) diagnosis of AD is done by observing markers such as neuro-psychological markers and markers found in structural imaging. Advances in early detection of AD are summarized in \cite{nestor2004advances}, \cite{swainson2001early} and \cite{dubois2009early}. Also there are some work specific on non-invasive diagnosis of AD, which can be found in \cite{sandson1996noninvasive}. In the literature of AD diagnosis, some works are concerned on biological and molecular studies (\cite{ray2007classification} and \cite{rojo2008neuroinflammation}) which these areas are not a concern in this paper. Some other works are focused on using non biological data (in non-invasive cases) such as facial emotion (\cite{spoletini2008facial}, \cite{burnham2004recognition}), handwriting (\cite{schroter2003kinematic}), spoken language, etc. In these areas there is a data provided from cases of AD and control cases, with the label specifying the class that each individual belongs to, then the task is to find a specific pattern in which it can be used to separate each classes members from the other.
	
	One of the most important markers in AD is language impairment which can be considered as a way of diagnosis in non-invasive manner. There are four main features to study in AD cases including semantic impairment, acoustic abnormality, syntactic and information impairment mainly discussed in \cite{fraser2016linguistic}.  Among methods that try to diagnose AD via spoken language, ones using machine learning had become very popular in recent years. 

	As an overview on works concentrated on designing a classifier to detect AD given the narrative speech followings are highlighted.  Warnita et al.\cite{warnita2018detecting}  used gated convolutional neural network on a set of emotion, paralinguistic, speaker state and speaker trait features derived from audio data and Wankerl et al.\cite{wankerl2017n} made their classifiers based on $N$-gram model and used it on text scribe. Zhu et al. had created neural network to learn low-dimensional representations \cite{zhu2018isolating}. Campbell et al had used SVM on data with GMM supervector kernel\cite{balagopalan2018effect}.  A brief overall view on these works are given in table \ref{tbl:WorkPrev}
	
	\begin{table*}[t]\label{tbl:WorkPrev}
		\caption{A Briefing on Previous Works}
		\centering
		\begin{tabular}{|c||p{0.45\textwidth}c|}
			\hline
			\textbf{Work} 	& \centering{\textbf{Method}} 	& \textbf{Achievement} \\ \hline \hline 
			\textbf{Warnita et al.} \cite{warnita2018detecting}	& Multilingual word embeddings, Clustering and Feature Extraction & 63\% (English) \\ \hline
			
			\textbf{Fraser et al.} \cite{fraser2016linguistic}	& Linguistic Feature Extraction, exploratory factor analysis & 81\%  \\ \hline
			
			\textbf{Wanker et al.} \cite{wankerl2017n}& N-gram model perplexity evaluation & 77.1\% \\ \hline
			
			\textbf{Zhu et al.} \cite{zhu2018isolating} & Neural network classifiers that learn low-dimensional representations
			reflecting the impacts of dementia yet discarding the effects of age & 76\% \\ \hline
			
			\textbf{Campbell et al.} \cite{balagopalan2018effect} & SVM with GMM Supervector Kernel
			&  \\	
			
			\hline
		\end{tabular}
	\end{table*}
	
	 Also there are some other studies using machine learning on the same data-set which we have used for evaluation including , \cite{zhu2018isolating}, \cite{balagopalan2018effect} and \cite{masrani2017detecting}.
	
	The idea of using machine learning on non-biological data to predict a type of dementia is not something new. Other similar works are done on other types of dementia such as Tahir et al. that used ANN and SVM on gait patterns \cite{tahir2012parkinson} or Jarrold et al. which made a classifier on lexical and acoustic features using logistic regression, multilayer perceptron and decision tree \cite{jarrold2014aided}. Also, Thomas et al. used ML algorithms on speech data to detect types of dementia \cite{thomas2005automatic} on ACADIE database \cite{rockwood2002goal}.
	
	In this paper a novel framework based on both acoustic and linguistic features of spoken language has been developed which involves both statistical and neural feature embedding techniques and perplexity evaluation. The remainder of this work organized as follows:
	
	Section \ref{sec:Proposed Framework} includes both an overview on our purposed framework and a briefing introduction to each building block and the theory behind it.  It is followed by section \ref{sec:Experimental Setup} including both an overview of the framework in applied manner and descriptions over evaluation data and experimental results. Conclusions are also being discussed in \ref{sec:Conclusion and Future Works}.


	\section{Proposed Framework}\label{sec:Proposed Framework}
	
	To diagnose AD, it is shown that both semantic context and acoustic features of speech data is appropriate to be used. Given that fact, we use a combination of models working with both the transcription of the speech and the audio format of the data. Recent studies have used syntactical and acoustics features of a speech data to predict Alzheimer's separately. In this work we are concentrated to use both of these representations to reach a better accuracy. 
	
	The system is supposed to receive an audio file from the spoken language as input and return a classification of whether the voice data is correspondent to an AD case or not. Usually the transcription of a speech data is not provided explicitly. In that situation an ASR (Automatic Speech Recognition) is used to extract a transcription from the audio file. In this study we have purposed three different models to be combined and to be used for the prediction process which are N-gram, I-vector and X-vector. N-gram is only used with the transcription of the spoken language as input while the other two models work with the acoustic features.

\begin{figure*}[h]
		\label{fig: procedure}
		\centering
		\includegraphics[scale=0.35]{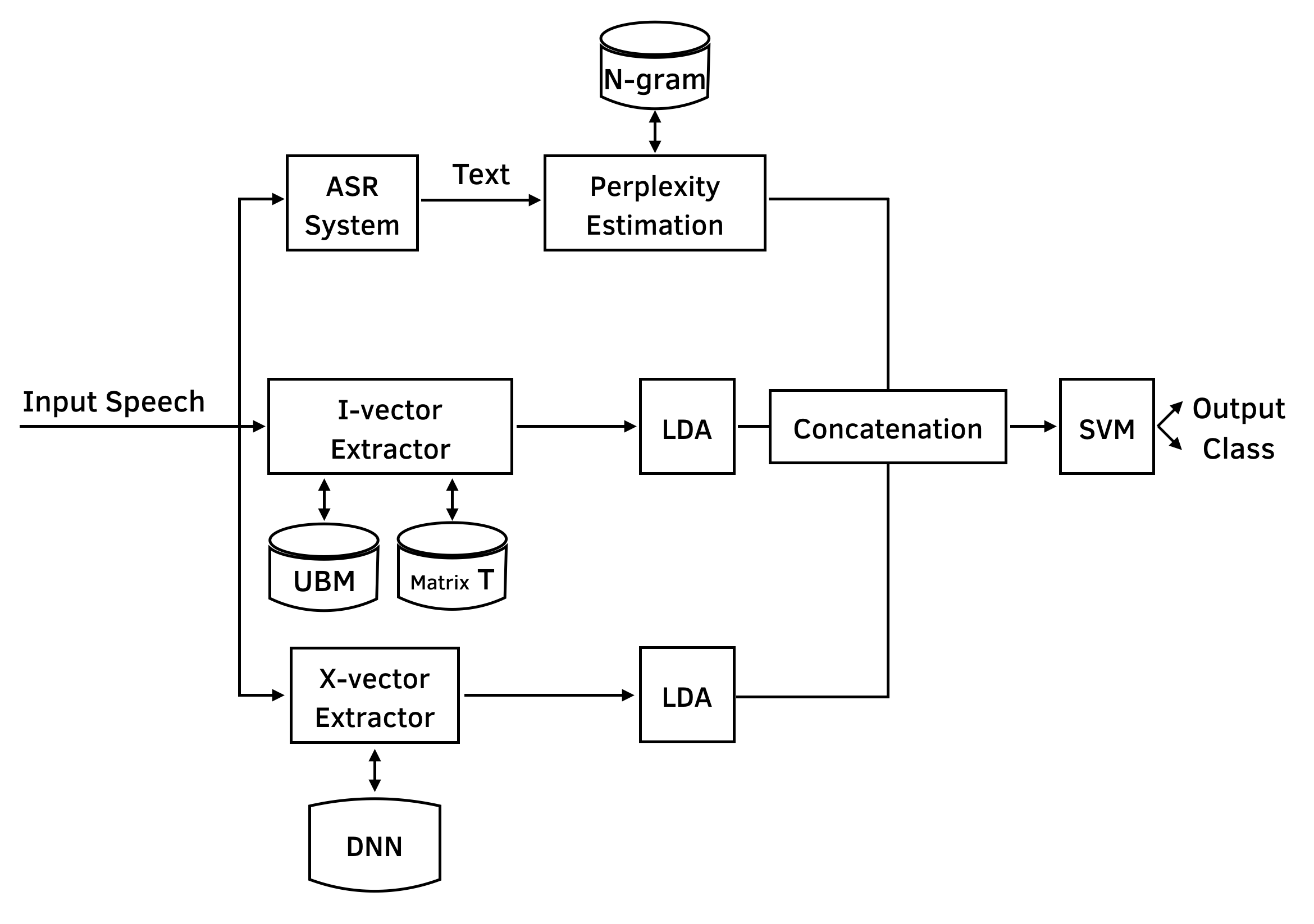}
		\caption{Block Diagram of the Purposed Framework}
	\end{figure*}

	As shown by figure \ref{fig: procedure}, the speech data which is represented as an audio file, will be processed by three models in parallel. Unless two models which receive audio representation of speech data (i-vector and x-vector), N-gram model, requires a transcription text which will be provided by ASR used as preprocessing. The result of N-gram model is a single scaler known as \textit{perplexity} for each class (dementia and control). Other two models return a vector which is an embedding of the received audio file. At last, the values returned by all these three models are concatenated into a single vector which is a representative of the received input and will be used for the classification task. For the classification task, a support vector machine is used which receives the resulting input and returns the prediction result.
	
	The N-gram model which receives a transcription text, is a model which stores a perception of consecutive usage of words in a corpus, in form of a set of conditional probabilities. By which a new sentence could be evaluated by a number, denoting how far is the sentence from the stored structure. The notation used for this distance is \textit{perplexity}. The less the perplexity, the more the sentence is supposed to be derived from the pattern of sequential occurrence of words in the corpus. In our study there are two N-gram models, trained one given the transcripts from control group and the other given the AD cases. A comparison of perplexities, shows how similar a case is to AD or control group. The setting of this model in this study is very similar to \cite{wankerl2017n}. 
	
	In addition to transcription and the sequential usage pattern of words, acoustic features of speech can have significant impact on the accuracy of diagnosis. As mentioned previously there are some impairments expected in both linguistics and speaking pattern of an AD case. Hence, we have involved two models of signal embeddings to track issues in the audio signal and use them to classify the cases of this problem. 
	
	In this study, two models (i-vector and x-vector) are used to extract features from the audio data. Both models are known from the literature of speaker identification as an embedding tool (\cite{dehak2011front} and \cite{synder2018spoken}). I-vector is a statistical model and x-vector is a model based on deep neural network. These two models are used to extract a feature vector from variable length audio signal. In addition to speaker identification tasks, these models are used in other areas such as language identification \cite{martinez2011language}, emotion recognition \cite{el2011survey}, music genre classification \cite{dai2017multilingual}, and online signature verification \cite{zeinali2018online}.

	Despite the fact that AD diagnosis and speaker recognition are different tasks, voice biometrics and Alzheimer signs are similar in nature as both need to extract some specific patterns from captured signal contaminated with variations from various irrelevant sources. As a result we expect that i-vector and x-vector should be able to provide a promising solution to the Alzheimer symptom extraction problem through spoken language.
	
%
	
	Following subsections will give a brief introduction to each method, and the theory behind.  
	
	\subsection{$N$-gram Model}\label{subsec:N-gram Model}
	Given an stream $s$ which in case of this work is a sequence of words ($s = s_1 , ... , s_n$), we seek to find a sequence of words $w$ which are the most predicted after the given. By the most predicated sequence we mean a sequence maximizing the maximum posteriori \cite{brown1992class}. 
	
	The $N$-gram model is a probability distribution over all possible sequences of words. The probability assigned to a sequence of words $S = (w_1 , ... , w_k)$ given the previous $n$ words is defined as \begin{align}
		\Pr(S) = \prod_{i = 1}^{k}\Pr(w_i\mid w_{i - n + 1} , ... , w_{i - 1})
	\end{align}
	Respecting to $N$, the model may become more complex. As an example for $N = 1$ in which the model is called uni-gram the only calculated probability for each sequence is $\Pr(S) = \prod_{i = 1}^{k}\Pr(w_i)$ and by increasing $N$ (to bi-gram tri-gram and generally $N$-gram) the model calculates $|\Sigma|^N$ probability function values, where $|\Sigma|$ is the cardinality of the set of all words in the language. 
	
	There are two main considerations in creating the model. First is that the model for $N > 1$ may be ambiguous for the elements at the beginning of the sentence; Other is that given any training data, there may be some usual sequences not mentioned in, which means that the calculated probability is zero. This issue may cause problems in evaluating a string with the model. To overcome these two problems, first we add an artificial token in the beginning of the sentence. Also to handle the second consideration,  we perform an smoothing on out model to make the probability of words or sentences that had not occurred in the input, some constant greater than zero. There are various ways to smooth the model, the simplest is adding a constant value to every $N$-gram result. Various smoothing techniques including one-count, average-count and etc are discussed in \cite{chen1999empirical}. 
	
	There are many smoothing procedures but in this work we used two of them called ``Good-Turing'' and ``Knezer-Ney''. In Good-Turing technique, we reallocate the probability mass function of those with $r+1$ occurrence to elements with $r$ occurrence. In particular we reallocate $N$-grams with one time expression in the data to ones never occurred. More accurately, we define a new count function $r^*$ as for each $r$ as a count \begin{align*}
		r^* = \frac{n_{r+1}}{n_r}(r+1)
	\end{align*} in which $n_r$ is the number of $N$-grams counted exactly $r$ times. By this, the new probability mass function of each $N$-gram $X$ is \begin{align*}
		\Pr(X) = \frac{r^*}{\sum_{i = 0}^{\infty}r^*n_r}
	\end{align*}. By the way there are some problems for Good Turing model as long as if for a model $n_{r+1} = 0$ then there will be problem. This case is called black hole.
	
	To evaluate a given stream of words $s= (w_1...w_k)$ which is calculation of that how much does the model fits a data, the \textit{perplexity} is defined as \begin{align}
		PPL(S) = \Pr(S)^{-\frac{1}{k}}
	\end{align} This definition also gives a weighted average branching in which the lower the perplexity is the less successors do the chain of words have. 
	
	An important concern in this model is time and space as it grows exponentially with respect to $N$ (the dependence level of $N$-gram model). This is the reason than models with large $N$ are not often used. Usually used models are bi-gram and tri-gram. 
	
	\begin{figure}
		\centering
		\includegraphics[scale=0.4]{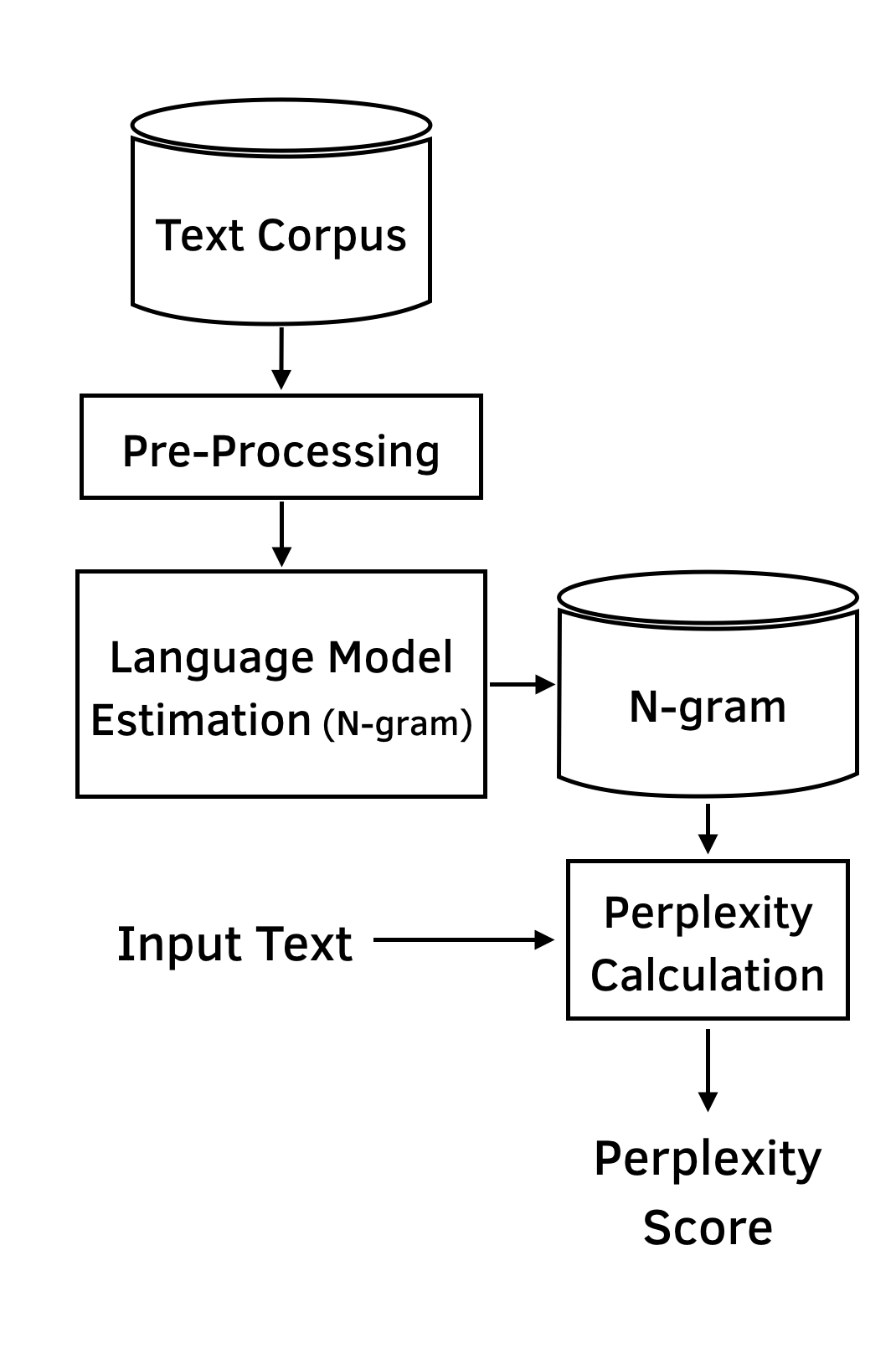}
		\caption{Block diagram for $N$-gram model.}
	\end{figure}

	\subsection{i-vector Approach}\label{subsec:i-vector Approach}
	This method has become the state-of-the-art in total variability space for speaker recognition \cite{dehak2011front}. Before this, commonly used method to solve speaker recognition tasks  was joint factor analysis \cite{kenny2007joint}. Also this approach is been used in many other stream processing task such as signature verification \cite{zeinali2017text} and ...
	Given an audio signal with arbitrary length, as soon as it's not represented in a fixed size vector and may appear in different length for each data element, using classifiers such as SVM or etc is unattainable. What i-vector does is to extract a fixed-length and compact vector representation from the audio (generally stream) and then we are able to use usual models of analysis, such as vector distance based similarity measures. Extracting i-vector from a given signal is done as following:
	\begin{itemize}
		\item \textbf{Universal Background Model:} First step for extracting i-vector is to create a background model. A Gaussian Mixture Model has been the most successful one for analyzing text-independent speaker recognition.\cite{reynolds2000speaker} Also for 
		text-dependent speech, methods like HMM are often used.\cite{zeinali2015telephony ,zeinali2016vector , zeinali2017text} Here as for each individual speaker, speech data elements may differ from one to one,  A GMM with $k$ multivariate Gaussian distributions is formulated as  
		\begin{align}
		\Pr(x\mid \lambda) = \sum_{i = i}^{k} w_i \mathcal{N}(x\mid m_i , \Sigma_i)
		\end{align}
		where $w_i$ denotes the weight given to $i$th component of mixture, $\mathcal{N}(x\mid m_i , \Sigma_i)$ is a Gaussian distribution with mean $m_i$ and diagonal covariance matrix $\Sigma_i$. Note that $\sum_{i = 1}^{k}w_i = 1$. Same as this study, usually GMM is used with a diagonal covariance matrix \cite{reynolds2000speaker}. 
		
		\item \textbf{Baum-Welch Statistics Extraction:} 
		In this step, having the universal model trained, and $X_i$ as the entire selection of feature vector corresponding to the $i$th data element, we can compute zero and 1st order of Baum-Welch statistics \cite{kenny2005eigenvoice}. We will calculate zero order BW statistics for the $j$th component of UBM as
		\begin{align}
		N_j = \sum_t \gamma^j _{i ,t}
		\end{align}
		and first order as
		\begin{align}
		F_j = \sum_{t} \gamma^j _{i , t} (X_{i , t} - m_j)
		\end{align}
		where $\gamma^j _{i , t}$ is posterior probability of generating $X_{i,t}$ by the $j$th component and calculated as
		\begin{align}
		\gamma^j _{i , t} = \Pr (j | X_{i, t}) =  \frac{w_j \mathcal{N}(X_{i , t}\mid m_j , \Sigma_j}{\sum_{s = 1}{k} w_s \mathcal{N}(X_{i , t}\mid m_s , \Sigma_s)}
		\end{align}
		\item \textbf{i-Vector:} 
		Construct $M$ as the individual dependent mean-supervector representing feature vectors of the sound, by concatenation of all $k$ mean vectors of the GMM for given data element. This \textit{super-vector} is modeled as
		\begin{align}
		\label{ivec}
		M = m + Tw
		\end{align}
		
		where $m$ is an individual independent super-vector derived from the UBM, $T$ is a low rank matrix, and $w$ is a random latent variable having a standard normal distribution. Also by the notation \textit{super-vector} we mean a $D.C$-dimensional vector made by concatenation of $D$-dimensional mean vectors of GMM corresponding to the data element. The i-vector $\phi$ is the MAP (maximum a posteriori) point estimate of the variable $w$ which is equal to the mean of the posterior probability of $w$ given the input data element. It is assumed that $M$ has a Gaussian distribution with mean $m$ and covariance matrix $T\times T^t$
	\end{itemize}
	\subsubsection{Training the Model}
	In the definition of i-vector (\ref{ivec}), parameters are $m$ and $T$. Usually $m$ is chosen as mean super-vector (concatenation of means of the UBM Components) of the universal background model \cite{campbell2006svm}. 
	For a UBM with $C$ components and $D$ dimensional feature vectors, matrix $\Sigma$ is formed as following
	\begin{align}
	\begin{bmatrix}
	\Sigma_{1} & 0 & 0 & \dots  & 0 \\
	0 & \Sigma_{2} & 0 & \dots  & 0 \\
	\vdots & \vdots & \vdots & \ddots & \vdots \\
	0 & 0 & 0 & \dots  & \Sigma_{C}
	\end{bmatrix}
	\end{align}
	where $\Sigma_c$ is the covariance matrix of the $c$th component of UBM.

	\begin{figure*}[h]
		\centering
		\includegraphics[scale=0.35]{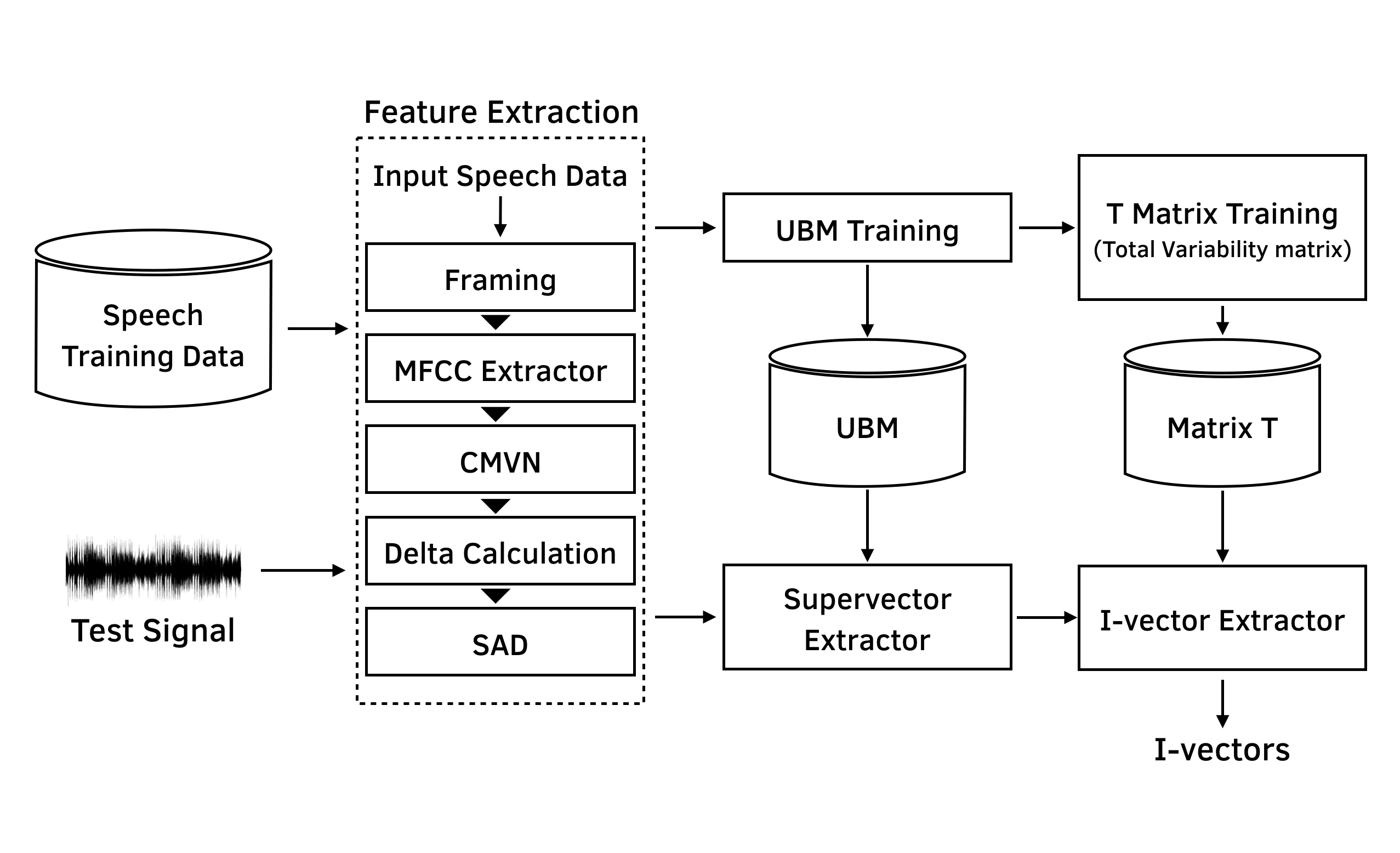}
		\caption{Block diagram for i-vector model.}
	\end{figure*}
	
	\subsection{X-Vector Approach}\label{subsec:x-vector Approach}
	
	One other approach of embedding audio used in speaker recognition tasks, is x-vector which has the same functional use as i-vector, since they both are tools to extract fixed length feature vectors from a variable length signal \cite{snyder2018x}\cite{snyder2018spoken}. Unlike i-vector which is said to be a statistical method, x-vector is based on feed forward deep neural network (DNN). It is notable that DNNs themselves can be directly optimized for speaker recognition tasks and there are some powerful and compact system designs based on them such as \cite{heigold2016end}\cite{variani2014deep}\cite{lei2014novel}. Even setting aside the speaker verification and recognition tasks, DNNs and its variations have a wide use in the literature of speech recognition \cite{hinton2012deep} \cite{deng2013new}, activity recognition \cite{zhang2015human} \cite{WANG20193}, handwritten recognition \cite{7280516}, etc.  The standard DNN, used for x-vector embedding, is based on the model presented in \cite{snyder2017deep} which has investigated an alternative for i-vector in text-independent speaker verification tasks. The DNN structure which we used in our setup, which is a little different with the original one, is outlined in table \ref{tbl:xvec conf}. 
	
	\begin{table*}[t]\label{tbl:xvec conf}
		\caption{Standard DNN Configuration for x-vector}
		\centering
		
		\begin{tabular}{l|ccc}
			\textbf{Layer} & \textbf{Layer Context} & \textbf{Tot. Context} & \textbf{In $\times$ Out}  \\ \hline\hline
			\textbf{Frame 1} & $[t - 4 , t+4]$ & 9 & $9F\times 128$  \\
			\textbf{Frame 2} & $\{t - 4 , t , t+4\}$ & 9 & $384\times 128$  \\
			\textbf{Frame 3} & $\{t - 5 , t , t+5\}$ & 15 & $384\times 128$  \\
			\textbf{Frame 4} & $\{t\}$ & 15 & $128\times 128$  \\
			\textbf{Frame 5} & $\{t\}$ & 15 & $128\times 7500$  \\ \hline
			\textbf{Stats Pooling} & $[0 , T)$ & $T$ & $7500T\times 15000$  \\
			\textbf{Segment 6} & $\{0\}$ & $T$ & $15000T\times 128$  \\
			\textbf{Segment 7} & $\{0\}$ & $T$ & $128\times 128$  \\
			\textbf{Softmax} & $\{0\}$ & $T$ & $128\times L$  \\
			
		\end{tabular}
		
	\end{table*}
	
	As table \ref{tbl:xvec conf} shows there are five frame stage layers, which are operating on speech frames with small temporal context from neighbor frames and each performing on the result of previous layer. The network receives $T$ speech frames, as a sequence. Each frame is supposed to be $F$ dimensional. As a result each frame stage layer receives a frame with a small temporal context (like 4th frame before and after for frame layer number 2) from the previous layer (1st frame layer receives the input) and leads to the result in 7500 dimensional vectors toward the statistics pooling layer. 
	
	The statistic pooling layer aggregates the result of previous layers across the time dimension, so that the result is an operation on the entire segment. This layer outputs the mean and standard deviation of each feature (dimension) over all frames (samples during the time). The output of this layer is the statistics (mean and std) over all features, concatenated on a single 15000-dimensional vector.
	
	Two remaining layer sets in the network are segment layers and Softmax. Setting aside Softmax, segment layers receive pooling layer's statistics and with non-linearity (which in our setup and the original one is set to ReLU), feed toward the final Softmax layer. We extract the embedding result from the layer ``\textit{Segment 6}'' before the non-linearity. Following figure shows the structure of DNN used in the work.
	
	Also it is important to note that as the end-to-end approach needs a large amount of in-domain data to work effectively, the loss function is multi-class cross entropy objective is used instead of end-to-end loss function.
	
	In this work we used Kaldi \cite{povey2011kaldi} as our toolkit which involves implementations of speech recognition algorithms and networks. Following figure shows a block diagram of the network, used in our work and further subsections will go through details of learning and embedding.

	\begin{figure}
		\centering
		\includegraphics[scale=0.4]{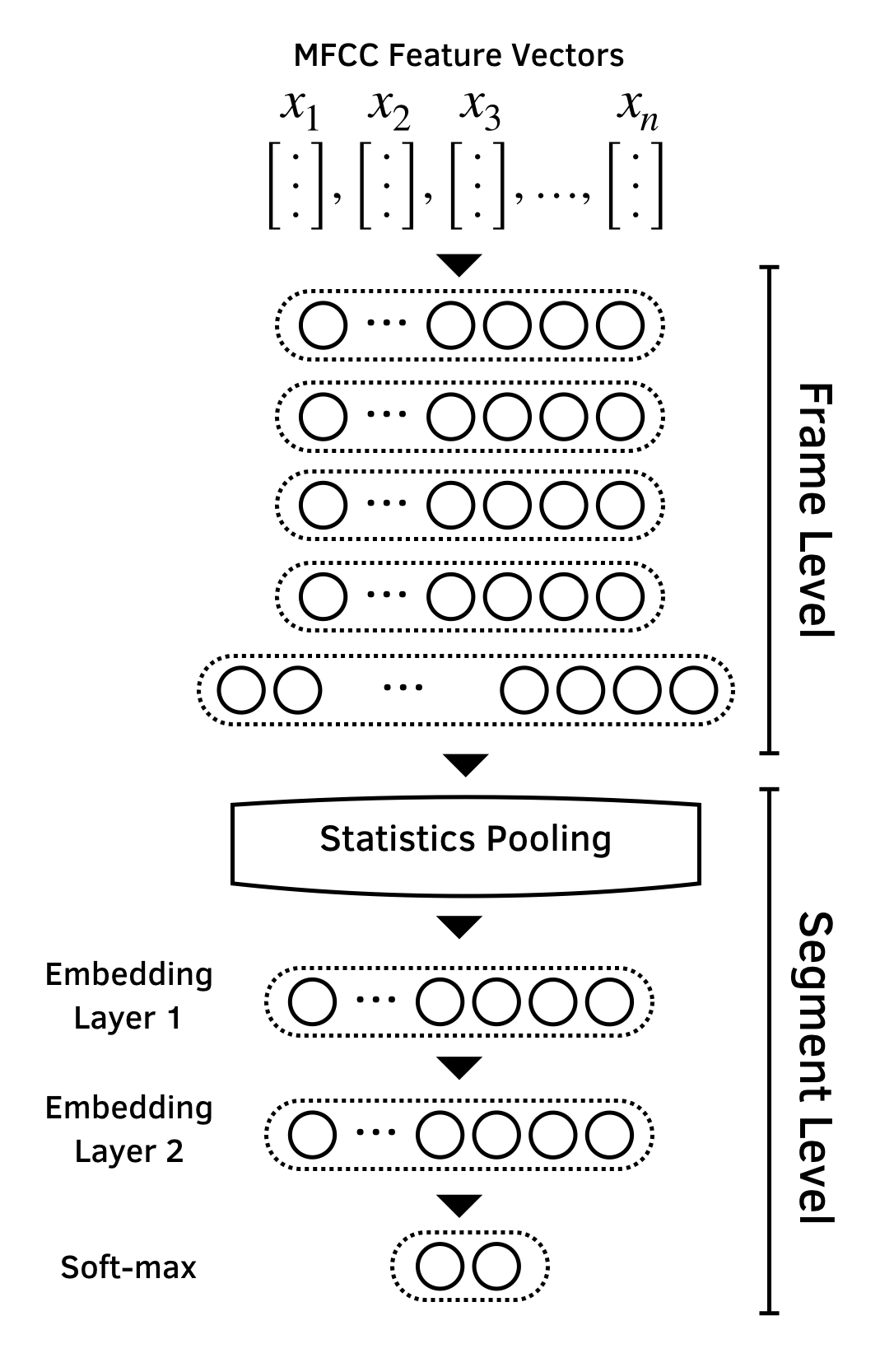}
		\caption{X-vector block diagram.}
	\end{figure}

	\subsubsection{Training the Model}
	The model is trained to classify the input sequences using multi-class cross entropy objective function. Given $K$ speakers in $N$ training segment $\Pr(s|x_{1:T}^{(n)})$ is the probability of speaker (class) $s$ given the stream $x_1^{(n)}, x_2^{(n)} , ..., x_T^{(n)}$, also $d_{n , s}$ is 1 if the speaker label for segment $n$ is $s$, otherwise 0. The mentioned objective function is \begin{align}
		E = - \sum_{n = 1}^{N}\sum_{s = 1}^{k}d_{n , s}\ln(\Pr(s|x_{1:T}^{(n)}))
	\end{align}. Training examples are taken from the training data as audio parts with duration over 2 to 4 seconds. The training procedure is done for several epochs using stochastic natural gradient descent.\cite{povey2014parallel}
	Also in training process, in order to increase the diversity of the data which is used to train the model, using data augmentation techniques is something common. One possible strategy is 6-fold augmentation which takes a clean training segment and creates 5 other copies with performing speed perturbation, music, noise and reverb randomly on its copies.
	
	\subsubsection{Embedding}
	The main goal of using both i-vector and x-vector models is to find an embedding from the variable length signal representation to a fixed length vector. Also there are many speaker recognition tasks done by frame level analysis but this method is supposed to find a feature vector correspondent to the whole signal without considering the length to be a fixed value. At test time, 512-dimensional x-vectors are extracted at layer \textit{segment6} of the network, before the non-linearity. Even the model is used as a mean of embedding and extracting constant sized feature vectors, it can be used as a classifier too.  

	\section{Experimental Setup} \label{sec:Experimental Setup}
	
	As a briefing to our model which is illustrated in figure \ref{fig: procedure}, the model receives an audio data as input and through a ASR system produces a transcript file which is an input for one of three lines of learning which involves $N$-gram model. Before evaluating whole framework, we examined each of the concurrent lines of it, evaluated them, then used the best configuration in our framework. Also at the end we tried every possible combination of lines via turning each of them on or off. Results will be discussed in subsection \ref{sec:Experiments and Results}. 
	
	During the evaluation of each model, for $N$-gram we examined bi-gram, tri-gram and 4-gram model with Good-Turing and Knezer-Ney smoothing function and used the most accurate one in our framework. Also for i-vector we made experiment to determine best combination of UBM components and i-vector size.

	\subsection{Data}\label{subsec:Data}
	In this study, we used Pitt Corpus \cite{becker1994natural}, the cookie picture description task, which is (to the preparation date of this work) including audio data and transcripts corresponding to each audio from 243 experiments on control (healthy) subjects and 309 experiments on patients having dementia including AD and probable AD. The cookie task involves an experiment in which an specific picture is shown to the interviewee and asks to describe the picture. Also alongside the recorded audio file, each file has a scribed text. 
	
	As data amount was limited, we performed 10-fold cross validation scheme. The test environment can be considered same as \cite{warnita2018detecting}, But despite Warnita et al. we performed our model on entire recordings of the data bank and did not ignore audio files with overlap from other interviews, which can be considered as fault tolerance in our model. 

	\subsection{Results}\label{sec:Experiments and Results}

	As discussed before of all, we performed each of the techniques separately and evaluated them on the data, then we use combinations of them to reach a better accuracy. First model to evaluate was $N$-gram model. In this work we used 2, 3 and 4-gram model with both Good-Turing and Kneser-Ney smoothing methods. Table \ref{tbl:Ngram} shows the result of $N$-gram model evaluation. 
	
	\begin{table*}[]\label{tbl:Ngram}
		\caption{$N$-gram Model Evaluation}
		\centering
	
			\begin{tabular}{l|ccccc}
				\textbf{N-gram} & \multicolumn{1}{l}{\textbf{Smoothing Method}} & \multicolumn{1}{l}{\textbf{Accuracy}} & \multicolumn{1}{l}{\textbf{Precision}} & \multicolumn{1}{l}{\textbf{Recall}} & \multicolumn{1}{l}{\textbf{F1-Score}} \\ \hline
				\textbf{2-gram} & Good-Turing& 78.2 & 79.1 & 78.2  & 77.8 \\
				\textbf{2-gram} & Knezer-Ney                                      & 74.4 & 77                                     & 74.3                                & 74.4                                  \\
				\textbf{3-gram} & Good-Turing                                     & 77.8                                  & 78.9                                   & 77.8                                & 77.5                                  \\
				\textbf{3-gram} & Knezer-Ney                                      & 67.6                                  & 71.1                                   & 67.5                                & 67.1                                  \\
				\textbf{4-gram} & Good-Turing                                     & 77.2                                  & 78.4                                   & 77.2                                & 76.8                                  \\
				\textbf{4-gram} & Knezer-Ney                                      & 67.6                                  & 71.1                                   & 67.5                                & 67.1                                 
			\end{tabular}
	\end{table*}
	
	Also i-vector evaluation is done with various sizes of UBM and target vector size which its result is gathered in table \ref{tbl:ivec}
	
	\begin{table*}[]\label{tbl:ivec}
		\caption{I-vector Model Evaluation}
		\centering
		
		\begin{tabular}{cc|cccc}
			\textbf{UBM Components} & \textbf{I-vector Size} & \multicolumn{1}{l}{\textbf{Accuracy}} & \multicolumn{1}{l}{\textbf{Precision}} & \multicolumn{1}{l}{\textbf{Recall}} & \multicolumn{1}{l}{\textbf{F1-Score}} \\ \hline\hline 
			512 & 512 & 74.9 & 75.1 & 74.8  & 74.6 \\
			512 & 256 & 73.8&	74.3&	74.0&	73.5 \\
			512 & 128 & 72.4	&72.9	&72.4	&71.9 \\
			512 & 64 & 68.9	&68.7	&68.8	&68.1 \\  \hline\hline          
			256 & 512 & 75.0&	75.9&	75.1&	74.8 \\
			256 & 256 & 75.6&	76.0&	75.5&	75.1 \\
			256 & 128 & 71.9&	71.8&	72.0&	71.3 \\
			256 & 64 & 68.6&	68.6&	68.5&	68.1 \\  \hline\hline 
			128 & 512 & 75.9&	76.5&	76.0&	75.5 \\
			128 & 256 & 75.0&	75.8&	75.1&	74.8 \\
			128 & 128 & 71.6&	71.8&	71.4&	71.1 \\
			128 & 64  & 68.2&	68.2&	68.1&	67.6 \\  \hline\hline  
			64 & 512 &  69.0& 	69.7&	69.2&	68.5 \\
			64 & 256 &  74.1&	74.9&	74.1&	73.6 \\
			64 & 128 &  71.9&	72.4&	72.1&	71.6 \\
			64 & 64  &  69.0&	69.0&	69.0&	68.5 \\     
		\end{tabular}
	\end{table*}

	After all, three models, i-vector, N-gram and x-vector are used combined in different settings to enrich the accuracy which shown in table \ref{tbl:combined}.
	
	\begin{table*}[]\label{tbl:combined}
		\caption{I-vector Model Evaluation}
		\centering
		
		\begin{tabular}{ccc|cccc}
			
			\textbf{X-vector} &	\textbf{I-vector}	&\textbf{Perplexity}	&\textbf{Accuracy}	&\textbf{Precision}	&\textbf{Recall}	&\textbf{F1-Score}\\ \hline \hline
			Yes &No	&No	&75.1	&75.3	&75.1	&74.8\\
			No	&Yes	&No	&75.9	&76.5	&76.0	&75.5\\
			No	&No	&Yes	&78.2	&79.1	&78.2	&77.8\\
			Yes	&Yes	&No	&76.7	&78.3	&76.7	&76.1\\
			Yes	&No	&Yes	&81.1	&82.0	&81.1	&80.9\\
			No	&Yes	&Yes	&83.1	&84.0	&83.1	&83.0\\
			Yes	&Yes	&Yes	&\textbf{83.6}	&\textbf{84.3}	&\textbf{83.6}	&\textbf{83.4}\\

		\end{tabular}
	\end{table*}
	
	\section{Conclusion and Future Works}\label{sec:Conclusion and Future Works}
	In this work we used three methods, two of them on the sound signal data (i-vector and x-vector) and one on the sequence of words (N-gram model) for diagnosis of Alzheimer disease and we reaches to accuracy 83.6\%. This model can be applied on various languages and even low-resource ones. 
	
	One of our future works will be evaluating these techniques specially i-vector and x-vector on other biological data, also one other possible works will be giving a mathematical model on symptoms of dementia such as Alzheimer to analyze them more precisely.
	
	\section{Acknowledgments}
	The data for evaluation of our framework was provided by at the University of Pittsburgh School of Medicine.

\bibliography{main}
\bibliographystyle{ieeetr}
\end{document}